# A Robotic Dating Coaching System Leveraging Online Communities Posts

Sihyeon Jo
Department of Electrical and
Computer Engineering
Seoul National University
Seoul, South Korea

Donghwi Jung
Department of Civil and
Environmental Engineering
Seoul National University
Seoul, South Korea

Keonwoo Kim
Department of Electrical and
Computer Engineering
Seoul National University
Seoul, South Korea

Eun Gyo Joung
Technology Management,
Economics and Policy Program
Seoul National University
Seoul, South Korea

Giulia Nespoli
Department of Industrial Design
Seoul National University
Seoul, South Korea

Seungryong Yoo
Department of Electrical and
Computer Engineering
Seoul National University
Seoul, South Korea

Minseob So
Department of Industrial
Engineering
Seoul National University
Seoul, South Korea

Seung-Woo Seo
Department of Electrical and
Computer Engineering
Seoul National University
Seoul, South Korea

Seong-Woo Kim
Graduate School of Engineering
Practice
Seoul National University
Seoul, South Korea

*Abstract*—Can a robot be a personal dating coach? Even with the increasing amount of conversational data on the internet, the implementation of conversational robots remains a challenge. In particular, a detailed and professional counseling log is expensive and not publicly accessible. In this paper, we develop a robot dating coaching system leveraging corpus from online communities. We examine people's perceptions of the dating coaching robot with a dialogue module. 97 participants joined to have a conversation with the robot, and 30 of them evaluated the robot. The results indicate that participants thought the robot could become a dating coach while considering the robot is entertaining rather than helpful.

*Keywords—chatbot, Human-Computer Conversation, machine learning, online community, Short Text Conversation*

## I. INTRODUCTION

Love and marriage are important events in the life of each person. People can feel deep emotions in dating relationships. Previous studies have shown that some people have difficulties in dating relationships and that providing people with the proper information can help couples pursue healthy relationships [1], [2]. Given the private nature of dating and marriage, there are potential needs and demands for dating coaching robots that are easy to access and can lower the mental barriers of users. Besides, applying appropriate and diverse advice to a given question leveraging data and machine learning schemes can significantly enhance a user's satisfaction. We have noted previous research that robot counseling robots have the potential to better engage the user than human counselors [3], [4].

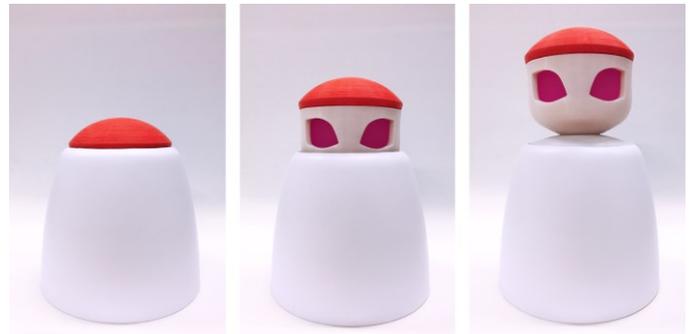

Fig. 1. DBO, a dating coaching robot.

In this study, we develop a Dating coaching BOt - DBO, a robot that can chat with users on the topic of love and marriage and give users advice. Fig. 1 shows the snapshots of DBO. We implement a data-driven chitchat engine. One of the most tricky parts to build the engine is to obtain explicit task-tailored semantics in the form of domain ontologies, hand-crafted by professional dating counselors. The lack of specialized conversational data makes it challenging to implement dialog engine modules in an end-to-end learning manner. However, casual dating counseling and guidance generated by collective intelligence are available in the online community. In this paper, we present a simple pipeline for implementing a dialogue module that exploits the common structure of posts and comments in the internet forums. Our experiment demonstrates that chatbots created along the pipeline have potentials to yield proper responses.

We propose a conversational robot capable of coaching a date with the introduction of design and implementation details. Using the 'thread-title and reply' structure of the online discussion forum as domain chatbot knowledge, we preprocess three college online communities posts on dating and relationships for using them as a user utterance-response (or Q-R) pairs for chatbot engine [5]. A dialogue engine is designed to enable conversation in the absence of dating counseling and coaching data provided by professional counselors. We extract common knowledge from datasets with different group personalities. There could hardly be a ground truth for a dating coaching conversation. Instead of quantitative evaluation using the test dataset, we evaluate the engine's performance and perceived characteristics of DBO in a user study.

## II. OUR PROPOSED SYSTEM

Previous studies have explored the potential of general counseling robots [6]. There were researches hypothesized that in the relationship between humans and robots, people would feel lower mental barriers and share more [3], [7]. Another work proposed positive effects of nodding for the robotic counseling system when listening to clients [4]. However, experiments of those studies were conducted under the limited use of conversation. We focus on the dialogue module because guaranteeing the quality of communication is a top priority in dating coaching scenarios.

Retrieval-based dialogue system conducts a dialogue by selecting an appropriate response for a given dialogue context, which differs from other conversation modeling paradigms such as generation-based methods [8]. The generation-based approach derives answers as a sequence of tokens generalizing over large-scale training data, with a tendency to produce general but non-specific answers such as "I don't know." [9]. Responses should be specific and controlled to prohibit inappropriate sentences. Therefore, we decide to build the dialogue module in the response selection manner.

We argue that we can extract chitchat knowledge from web forums for dating coaching with simple preprocessing. This argument is worth to be examined in that the post and replies pairs are similar to the chatbot template, and up-to-date information is posted on the dating counseling forum. Different opinions and expressions to the same question can be utilized as various responses to the same query in a response selection setting. We solve the Short Text Conversation (STC) problem, defined as a task that derives a proper response for the user's single query [10]. Formally, for a given query $q$, let $D$ be the given post-replies pair, which can be leveraged as chitchat knowledge. The retrieval-based short text conversation system retrieves an appropriate response based on the following three stages as presented in Fig. 2.

- Step 1 (Retrieval), in which the system retrieves $I$ that is a subset of $D$ based on $q$. Doc2vec embedding scheme is applied to reduce retrieval time with the dense representation of texts [11].

$$I = Retrieve(q, D). \quad (1)$$

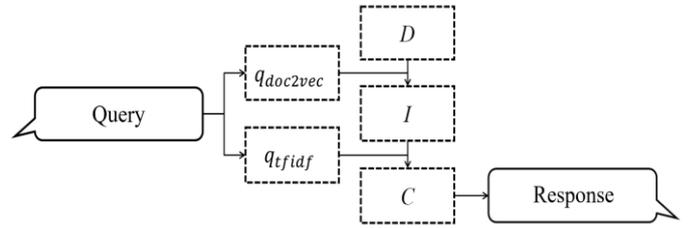

Fig. 2. The pipeline of the dialogue module.

- Step 2 (Matching), in which the system selects candidate responses $C$ from $I$ by comparing matching scores calculated between $q$ and $I$. TF-IDF embedding is adopted to include more relevant sentences into the response candidates set.

$$C = Match(q, I). \quad (2)$$

- Step 3 (Selecting), in which the system ranks all response candidate $r$ in $C$ and chooses proper response $\hat{r}$:

$$\hat{r} \in \underset{r \in C}{\mathrm{argmax}}\, Rank(q, r). \quad (3)$$

Let us assume $q$ is a $d$ dimensional vector and $r_i$ is the representation of the $i$-th candidate response in $C$ with dimension $d$. Let $W_j \in \mathbb{R}^{d \times d}$ and $b_j \in \mathbb{R}$ are the trainable parameters ($0 \leq j < m, j \in \mathbb{Z}, m \in \mathbb{N}$). Then, we can define the response selection model as follows:

$$f_{ij} = \sigma(q^T W_j r_i + b_j), \quad (4)$$

$$g_i = \sigma(f_i^T s + c), \quad (5)$$

$$p_i = softmax(g_i), \quad (6)$$

where $f_i \in \mathbb{R}^m$ represents the relational features between $q$ and $r_i$, and $g_i \in \mathbb{R}$ includes the information about score of the $i$-th candidate response. $\sigma$ denotes the activation function such as ReLU. Lastly, the $i$-th candidate response is selected with the possibility of $p_i$.

Using the Mean Squared Error (MSE) loss, $p_i$ is learned to have similar values with the $i$-th candidate response's score calculated by dividing the $i$-th *Likes-Dislikes* by total sum of *Likes-Dislikes* in $C$. The suggested model can be trained with an optimizer such as Stochastic Gradient Descent (SGD).

The response is chosen based on the value of $p_i$ resulting in a variety of interesting answers. Also, the proposed response selection model has the advantage that it can be applied to online community data having no user preference information on comments by learning to find a proper response for a given query among retrieved candidate response.

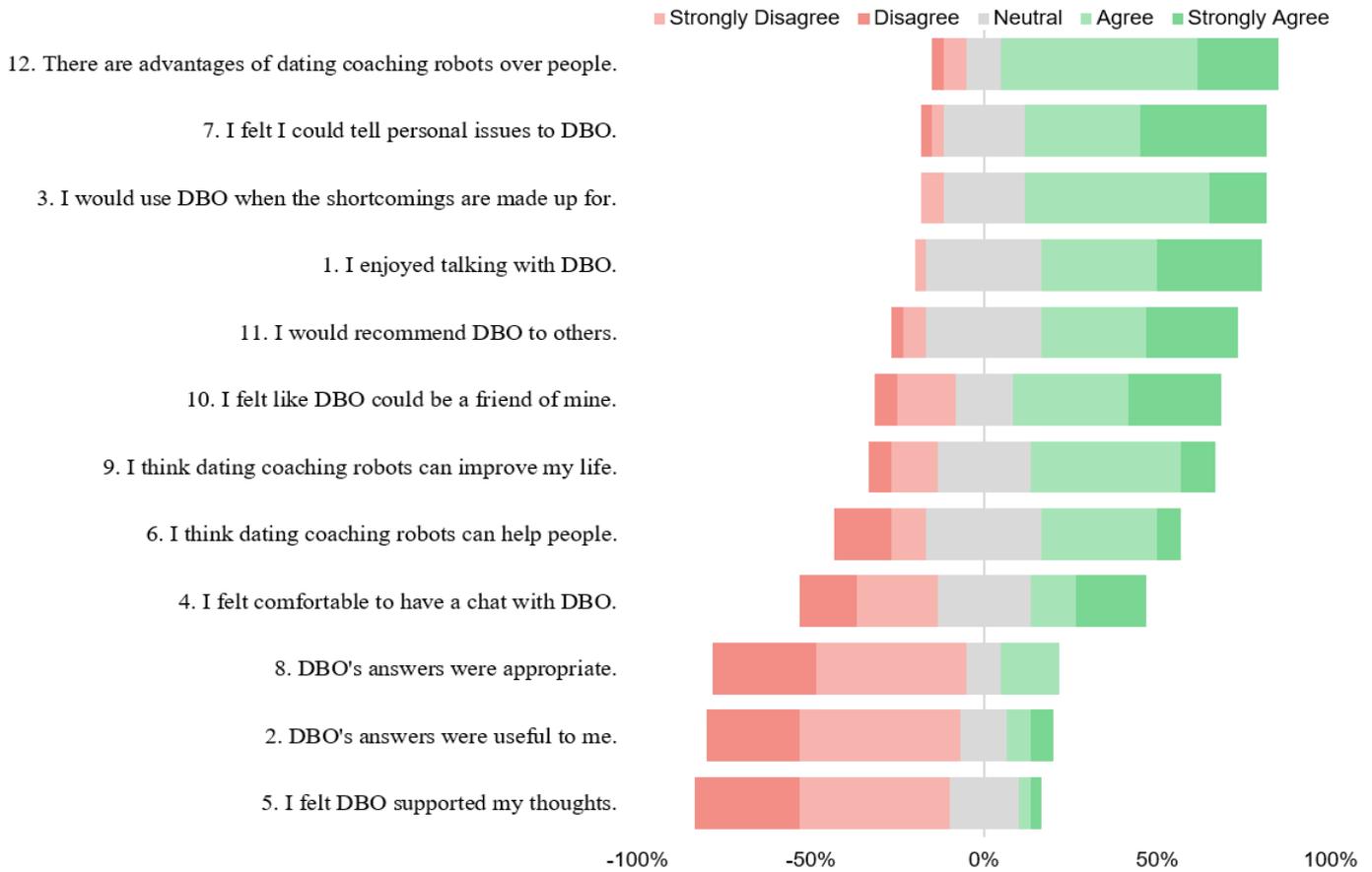

Fig. 3. Responses to each question. This shows the percentage of Likert response.

## III. EXPERIMENTAL EVALUATION

### A. Data Collections

We crawl post-comments pairs from college student online dating counseling forums named SNULife, Koreapas, Seiyon.net respectively. We remove non-text-based posts such as photos and explicit noises such as advertisements from the corpus. We also keep out the posts with no comments or posts with no main text. The number of posts satisfying these conditions is 267,789, and the number of replies is 2,829,797. Then, after tokenization with KoNLPY [12], titles are embedded in 256 dimension vector space with distributed memory doc2vec [13] model trained with the entire title corpus. Replies are embedded in 128 dimensions in the same way.

### B. Experimental Setup

We conduct experiments three times using an online messenger platform. Participants are allowed to have a conversation freely with DBO for up to two hours for each trial. With 97 participants, there are 238 questions, 549 chats, and 398 conversation rounds. There is the participant chatting with DBO up to 9 questions, 52 conversations, and 21 rounds at the maximum. At the median value, people ask 2 items, enjoy 4 chats, and experience 3 turns of conversation. Participants are asked to evaluate DBO willingly, and 30 of the counselee fill out forms.

### C. Results

Fig. 3 shows that the participants rate DBO as friendly and agree that there are comparative advantages in dating coaching robots over human counselors for some attributes such as privacy and accessibility. We conclude that DBO have the potential to better engage the counselee than human counselors with the results for question 12, 7, 3 and 1, which are more than 80% of the participants respond positively.

We observe that the participants find DBO entertaining rather than helpful. Question 5, which is the only item more than 80% of the participants answered negatively, shows that responding with expressions of sympathy can also be an issue to be solved in order to create a good impression to users. From this, we may interpret that people have high expectations for the usefulness and appropriateness of advice and responses from chatbots in dating coaching scenarios.

## IV. CONCLUSION

In this paper, we verified the possibility of constructing a data-driven dating coaching robot by processing posts from the online communities dataset and exploiting posts' structure to use them as chatbot knowledge. We evaluated our hypothesis on real-world dating coaching scenarios with 97 participants and obtained promising results for dating coaching robots' entertainment purposes.

ACKNOWLEDGMENT

This work was financially supported by the Ministry of Culture, Sports and Tourism, and the Korea Creative Content Agency, in part the National Research Foundation of Korea through the Ministry of Science and ICT under Grant 2018R1C1B5086557, and by Korean Ministry of Land, Infrastructure and Transport (MOLIT) as 「Innovative Talent Education Program for Smart City」.